# Action Representation for Spatial Generalization in Robot World Models

Yuhang Sun
University of Washington, Seattle, WA 98195, USA
*Correspondence:* sunyhg@uw.edu

## Abstract

Spatial generalization in robot world models depends on whether action representations capture the physical relationships that remain valid across changing spatial configurations. Yet different coordinates can describe the same interaction, whereas similar motions can produce different outcomes when local physical conditions change. How these distinctions affect learned prediction remains insufficiently understood. This study evaluates matched action representations defined in world, object, and surface coordinate frames using a controlled robotic peeling benchmark. Global pose and contact relocation experiments reveal a cost associated with limited training data when equivalent interactions receive different spatial descriptions. Removing information about peel thickness creates an irreducible ambiguity because identical predictive inputs can correspond to different material removal outcomes. Coordinates defined relative to the surface improve consequence prediction and counterfactual action selection under accurate local observations, but learned predictors remain vulnerable to uncertainty in the coordinate frame. Joint centering of the surface and tool removes sensitivity to translation of their common origin, whereas incorrect contact geometry and actual execution deviations introduce different failure mechanisms. These results show that effective spatial generalization depends on preserving the relationships between actions and states that determine physical consequences, rather than simply minimizing coordinate variation. This perspective provides a principled basis for designing robot world models that reuse physical knowledge across spatial configurations while retaining the distinctions essential to prediction and planning.

## 1. Introduction

Robot world models seek to predict how actions change the physical environment and to support planning beyond previously observed situations. An essential capability is spatial generalization, whereby experience acquired in one configuration remains useful when objects, tools, and contact locations change. This capability is often associated with richer scene representations, but representing the environment and representing an intervention are distinct problems. A model can reconstruct where an object is located without knowing which changes of action coordinates preserve the physical interaction or which local differences alter its outcome. The question addressed here is how the representation of an action governs the transfer of learned action–consequence relationships across space.

Research on world models has progressively expanded their predictive scope. Early approaches demonstrated that learned latent dynamics could support policy evolution, planning from images, and control through imagined trajectories [1-3]. DayDreamer extended world-model learning to physical robots [4], while DreamerV3 demonstrated effective world-model-based control across diverse tasks[5]. More recent work has enriched the structure of the predicted environment. DINO-WM learns dynamics in pretrained visual feature spaces[6], DreMa combines reconstructed scenes and simulation for robot imitation learning[7], and ReViWo investigates view-invariant dynamics for visual manipulation [8]. PointWorld further develops three-dimensional representations of robot actions and scene evolution [9]. These advances make physical and spatial prediction increasingly expressive. However, they do not uniquely prescribe the action coordinates through which learned physical relationships should be reused in new spatial configurations.

A parallel line of research has developed structured representations for manipulation. Object-centric learning separates interacting entities and their dynamics, providing an alternative to treating a scene as an undifferentiated observation [10-12]. Transporter Networks exploit spatial correspondences for rearrangement [13], and Equivariant Transporter Networks introduce rotational structure into the corresponding action predictions [14]. Neural Descriptor Fields and Equivariant Descriptor Fields use SE(3)-structured object representations to support manipulation across changing object poses [15, 16]. These studies demonstrate that suitable geometric structure can improve the organization and transfer of manipulation knowledge.

The same geometric perspective has increasingly informed action generation and policy learning. Diffusion-EDFs, EquiBot, ET-SEED, and Equivariant Diffusion Policy incorporate transformations of spatial structure into generative manipulation policies [17-20]. Related work on Canonical Policy, Eq.Bot, and observation–action canonicalization seeks to simplify learning by standardizing how spatial configurations are expressed [21-23]. Studies of deformation-based shape servoing, generic contact-rich task models, and geometric optimal control further emphasize the role of task-dependent descriptions of geometry and action [24-26]. These contributions establish strong reasons to exploit object structure, symmetry, and task frames. Nevertheless, a geometric transformation is useful for consequence prediction only to the extent that it respects the physical relationship being predicted. Theoretical work on correct and incorrect equivariance reinforces the risk of imposing symmetries that do not match the target relationship [27].

This distinction exposes a question that geometric canonicalization alone cannot settle. Two actions may appear different in world coordinates while generating the same physical consequence because their local interaction conditions are unchanged. Conversely, actions with similar coordinate descriptions may produce different consequences if local geometry, material state, or actual tool motion differs. A representation that treats the first pair as unrelated makes the predictor infer an otherwise shared relation from separate examples. A representation that treats the second pair as identical creates a more fundamental problem, because the resulting input cannot uniquely determine both outcomes. Thus, the appropriate equivalence between actions cannot be inferred from spatial similarity alone. It must be justified by the physical conditions and prediction target of the interaction.

A peeling interaction makes the distinction concrete. Suppose a blade removes a strip of peel from a fruit. If the fruit and blade undergo the same rigid translation or rotation while their relative blade–surface motion and all relevant physical conditions remain unchanged, the material removed relative to the fruit is unchanged. The new world-coordinate trajectory describes no new cutting law. A model that has learned this interaction should therefore be able to apply the same predictive relationship at the transformed pose. Making the equivalence explicit in the representation can reduce the number of spatial variations that a finite-data learner must infer. This is an advantage of a suitable inductive bias rather than a claim that a sufficiently trained world-coordinate predictor cannot learn the equivalence on its own.

Now consider a different change. The blade follows an identical local path and penetration depth over two patches with different peel thicknesses. One stroke can expose the flesh while the other leaves residual peel. If peel thickness, or any equivalent information about the local material state, is absent from the entire predictive input, the model receives indistinguishable inputs for different required outputs. No deterministic prediction function can be correct for both. Unlike the finite-data burden associated with redundant coordinates, this ambiguity cannot be removed merely by increasing model capacity or optimization effort. The necessary distinction may be carried by the state representation rather than by the action coordinates themselves, but it must remain available to the predictor.

These two cases lead to complementary requirements. Consequence-preserving invariance concerns spatial transformations that change the description of an interaction without changing its fruit-attached physical outcome. Encoding such a known equivalence can facilitate transfer of learned relationships across configurations. Predictive sufficiency concerns the joint state–action input and requires it to distinguish interactions whenever their target consequence distributions differ. The requirements are not interchangeable. Removing too little redundant spatial variation can make generalization unnecessarily data-intensive, whereas removing too much physical information can make accurate prediction impossible. They also depend on the prediction target. Fruit-attached material-removal measures may be invariant under a common rigid transformation even when the corresponding output geometry, expressed in world coordinates, must transform equivariantly. Changes in gravity-relative mechanics, obstacles, or other external conditions need not preserve the consequence and therefore cannot be assumed to be invariant.

A further gap separates physically justified representations from the behavior of learned predictors. A local coordinate frame can change numerically because of uncertainty in its estimated origin or orientation, even when the underlying blade–surface interaction remains fixed. A predictor using absolute local coordinates may respond to these changes despite the invariance of the true consequence. By contrast, selecting the wrong contact patch changes the physical information supplied to the model, and an error in the executed blade trajectory changes the actual intervention. These mechanisms should not be conflated under a single label of frame robustness. Distinguishing them is necessary to determine whether a failure of spatial generalization arises from redundant coordinates, lost predictive information, or a learned model that does not realize the relevant physical equivalence.

Robotic peeling offers a particularly discriminating setting for examining these issues because the consequence of a short cutting stroke depends jointly on contact location, local surface geometry, blade orientation, penetration, and material-layer thickness. Prior robotic peeling studies have investigated dual-arm coordination, multimodal and human-in-the-loop interaction, dexterous reorientation, and fine-grained knife control [28-31]. Complementary cutting simulators, including DiSECt and TopoCut, illustrate how tool–material interactions can be modeled for inference or policy learning [32, 33]. These developments establish the importance and complexity of peeling and cutting, but the present work uses the task for a different purpose. A controlled geometric material-removal simulator makes it possible to change the action representation while holding the underlying physical interaction and its target consequence fixed.

The study compares world-, object-, and surface-relative action descriptions using matched physical inputs, predictor architectures, training targets, and optimization. Common rigid transformations and surface-contact relocations test whether physically equivalent interactions can share learned predictions. Controlled thickness changes test the consequences of losing necessary state information. Additional perturbations separate estimated-frame changes from incorrect geometric observations and actual blade-execution deviations, while counterfactual action ranking connects predictive accuracy to decision quality. The analysis addresses prescribed short-horizon strokes and geometric removal outcomes rather than the full fracture, adhesion, sensing, or closed-loop force-control mechanics of robotic peeling. Within these defined limits, the contribution is a consequence-based account of action representation for spatial generalization. The results explaining which learned physical relationships can be transferred, which distinctions must remain available, and why representation-level equivalence may not automatically be realized by a trained world model.

## 2. Materials and Methods

### 2.1 Study setting and action-consequence prediction

A controlled test of action representation must separate differences in coordinates from differences in physical interventions. Broad manipulation suites provide complementary evaluations. RLBench spans varied robot tasks[34]. ManiSkill2 tests generalizable manipulation across rigid and soft-body settings[35]. Action-conditioned models have been studied in object pushing[36]. Such benchmarks involve many simultaneous sources of variability. The present setting instead isolates a small, observable physical consequence so that coordinate redundancy and the loss of necessary state information can be examined without conflating them with grasp planning or long-horizon control.

Each trial contains one fruit, a prescribed blade stroke, and a geometric material-removal model. A short-horizon interaction is used to isolate action representation from grasp selection, reorientation, and long-term force control.

Let its physical state be

$$X = (\Sigma, G, h, q). \quad (1)$$

Here $\Sigma$ is the fruit surface in a body-fixed fruit coordinate system, $G$ is its rigid pose in the world frame, $h(s)$ is the target peel thickness, and $q(s)$ is the accumulated removal depth at surface point $s$ on $\Sigma$. The fields $h$ and $q$ have units of length, and $h(s)$ is strictly positive. The state is evaluated immediately before each stroke. A candidate peeling action is parameterized by

$$\mathbf{A} = (\mathbf{p}_0, \psi, L, d, \alpha, w_b). \quad (2)$$

Here $\mathbf{p}_0$ is the initial contact point in fruit-body coordinates, $\psi$ the tangent direction angle, $L$ the prescribed stroke length, $d$ the nominal penetration depth, $\alpha$ the blade inclination angle, and $w_b$ the blade width. Angles are measured in radians and distances in meters. The blade width is fixed during each stroke. The variable $d$ describes commanded depth, whereas $q(s)$ denotes actual material removal. The simulator computes the incremental removal field

$$\Delta q(s) = q_{\text{post}}(s) - q_{\text{pre}}(s). \quad (3)$$

The subscript pre denotes the removal depth before the stroke, and post denotes the depth afterward. The coordinate $s$ identifies a point on the fruit surface. For the single-stroke experiments starting with intact peel, $q_{\text{pre}}(s) = 0$ and $q_{\text{post}}(s) = \Delta q(s)$.

The peeling outcome is summarized by $\mathbf{Y} = (C, U, O, V)$, where $C$ is coverage, $U$ is residual peel, $O$ is flesh loss from overcutting, and $V$ is depth nonuniformity. These values are computed from the complete removal field. The action-conditioned world model predicts this outcome from a chosen representation of the physical state and action

$$\hat{Y} = W_\theta\big(R(X, \mathbf{A})\big). \quad (4)$$

Here $R$ transforms the physical state and action into the selected input coordinates. the function $W$ with trainable parameters $\theta$ is the learned predictor, and $\hat{Y}$ denotes its estimated consequence. Each coordinate representation uses the same physical interaction and target outcome, allowing the effect of representation choice to be isolated.

## 2.2 Action representations and predictive requirements

Three descriptions of the same fruit–blade interaction are compared. World coordinates specify the blade and surface in a fixed workspace frame, so common movement of the entire interaction changes the input. Object-relative coordinates specify blade motion with respect to the fruit, removing changes attributable only to global fruit pose. Surface-relative coordinates additionally express the blade relative to the contacted surface patch. These are progressively more localized descriptions, not an unconditional ranking of quality. Their value depends on whether the physical conditions that determine material removal remain represented. All three baselines receive matched surface geometry, normals, and peel thickness; only their spatial coordinate descriptions differ.

Let $G$ denote the homogeneous rigid transform from the body-fixed fruit frame $B$ to the world frame $W$. The blade poses are indexed by a stroke parameter $\tau$, which identifies positions along the prescribed path rather than elapsed physical time. The superscripts $B$, $W$, and $S$ identify fruit-body, world, and local-surface coordinates. The subscript $b$ denotes the blade. The object-relative blade pose follows the standard rigid-transform relation

$$G_b^B(\tau) = G^{-1} G_b^W(\tau), \quad (5)$$

Let $g$ be a common rigid translation and rotation applied to the entire fruit–blade system. Its action cancels when the blade pose is expressed relative to the fruit

$$(gG)^{-1} g G_b^W(\tau) = G^{-1} G_b^W(\tau). \quad (6)$$

This identity establishes exact global-pose invariance for the object-centered action. It applies when the common transformation leaves all relevant external physical conditions unchanged.

At the initial contact point $\mathbf{p}_0$, the local frame is formed from the outward unit surface normal $\mathbf{n}(\mathbf{p}_0)$ and two mutually orthogonal unit tangent vectors $\mathbf{t}_1(\mathbf{p}_0)$ and $\mathbf{t}_2(\mathbf{p}_0)$

$$F(\mathbf{p}_0) = [\mathbf{t}_1(\mathbf{p}_0)\ \mathbf{t}_2(\mathbf{p}_0)\ \mathbf{n}(\mathbf{p}_0)]. \tag{7}$$

The local tangent directions are determined from a fixed object-axis reference and the surface normal, independently of the blade direction. The surface pose is a rigid transform with rotation $F(\mathbf{p}_0)$ and translation $\mathbf{p}_0$, both expressed in fruit-body coordinates. The blade pose relative to the surface is

$$G_b^S(\tau) = \left(G_s^B(p_0)\right)^{-1} G_b^B(\tau). \tag{8}$$

The surface input contains the sampled local point-normal geometry and peel thickness, together with the same physical action information supplied to the other baselines. It can make two contacts appear similar when the local surface and tool relationship really are similar. This is useful because the predictor need not learn the same local cutting relation separately at every object location. The claim is conditional: local shape, thickness, and blade motion must still distinguish contacts with different consequences, and exact relocation equivalence is generally restricted to controlled geometries.

The first requirement follows from the transfer question in the Introduction. Write $Z = (X, \mathbf{A})$ for the complete state–action interaction, and let $R(Z)$ denote its predictive representation. Suppose a transformation $T$ only changes the coordinate description of an interaction, while the relevant physical conditions and fruit-attached consequence stay fixed. A predictor could learn to give consistent outcomes for $R(Z)$ and $R(TZ)$, but an invariant representation makes that known equivalence explicit before training. Consequence-preserving invariance is expressed as

$$R(Z) = R(TZ). \tag{9}$$

Equation (9) removes a known, outcome-preserving coordinate distinction at the input level. It is a sufficient inductive bias for sharing examples across the corresponding transformations, not a logical claim that a noninvariant world-frame predictor must fail. A model may instead learn the relation or encode the appropriate equivariance. Its finite-sample cost is an empirical question. In this study the target is fruit-attached removal. If the predicted output were expressed in world coordinates, an appropriately transformed output would be required.

The second requirement follows from the opposite possibility. Two physically different interactions are mapped to exactly the same complete predictive input. When their consequences differ, no deterministic function of that input can predict both correctly. Thus the represented action together with the available state must retain the information needed for the target prediction. A representation is predictively sufficient if it retains all the information needed to predict the outcome:

$$P(Y \mid Z) = P\big(Y \mid R(Z)\big). \tag{10}$$

The condition does not demand a unique feature vector for every distinct physical scene. It demands only that scenes with different outcome distributions remain distinguishable for the prediction being made. Consider the thickness example: if peel thickness is absent from every predictive input channel, a thin-peel and a thick-peel stroke may have identical represented inputs but different outcomes. The resulting ambiguity can be quantified. In the deterministic simulator, let two equally weighted interactions have distinct normalized consequence vectors $y_1$ and $y_2$ but the same represented input $r$. Any deterministic prediction $\hat{y}(r)$ must then incur the following mean squared Euclidean error:

$$\frac{\|\hat{y}(r)-y_1\|^2+\|\hat{y}(r)-y_2\|^2}{2} \geq \frac{\|y_1-y_2\|^2}{4}. \tag{11}$$

Here $\|\cdot\|$ denotes Euclidean norm on a fixed, componentwise normalized outcome vector. The inequality holds for the equally weighted pair, independently of the architecture or training procedure. Its strictly positive lower bound, whenever $y_1 \neq y_2$, is caused by the indistinguishable input, not by a failure to optimize the network. Additional samples cannot reveal a variable that the predictive representation has discarded. This is fundamentally different from the possible finite-data cost of retaining unnecessary coordinates, which does not impose the same information-theoretic impossibility.

A third distinction concerns implementation. Even when the chosen frame expresses a physically justified equivalence, an unconstrained predictor may still respond to irrelevant changes in the numerical origin or orientation of its input. A common frame-origin shift, for example, translates both the surface points and blade positions without changing their pairwise relative geometry or the geometric-Oracle target. By contrast, selecting the wrong contact patch can remove necessary observations, and executing a different blade motion changes the target itself. These cases require different remedies and are analyzed separately. **Figure 1** illustrates the two foundational physical comparisons: common rigid motion preserves fruit-relative cutting consequences (**Figure 1** (a)), whereas different local peel thickness changes the consequence of the same blade motion (**Figure 1** (b)).

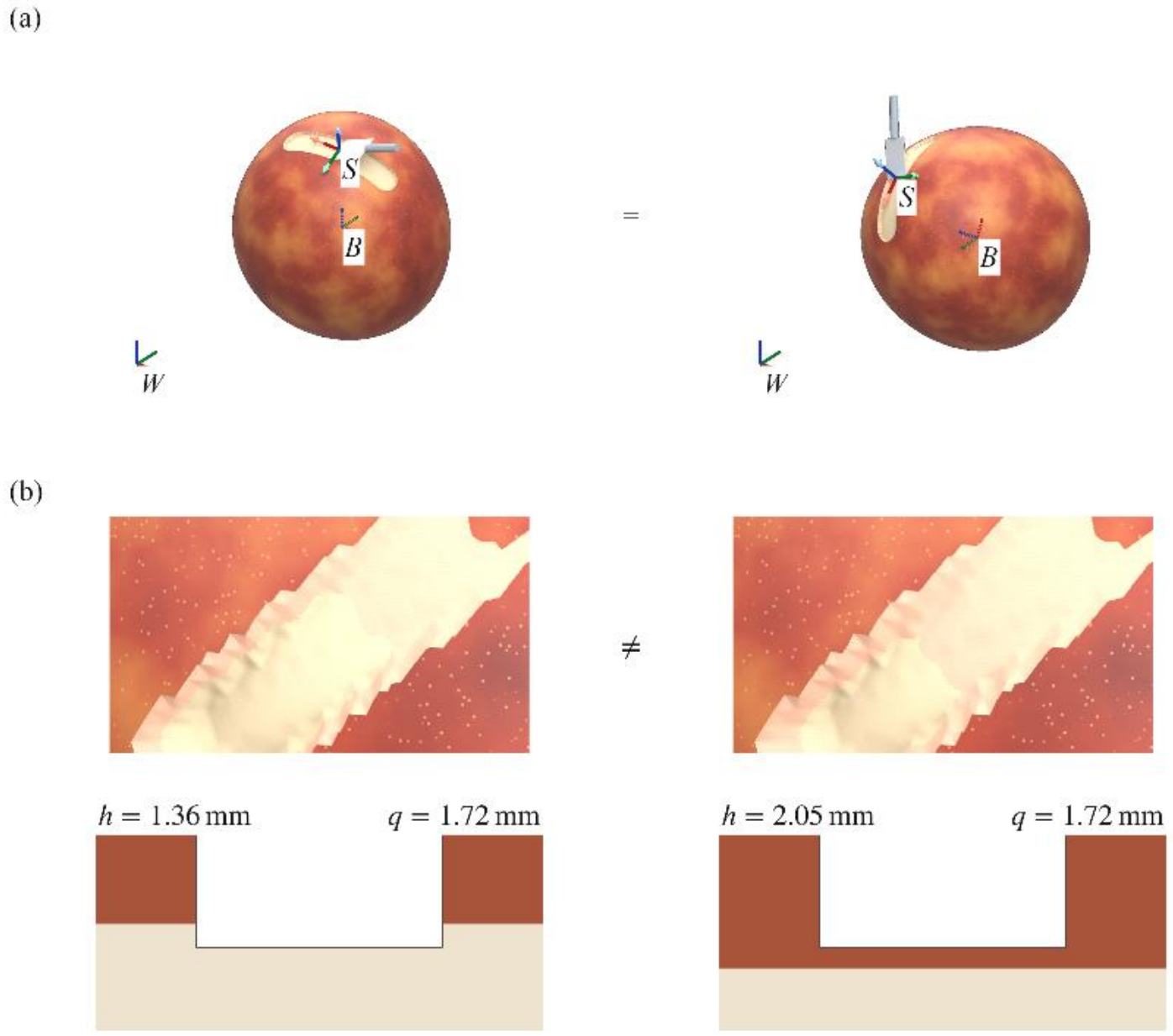

**Figure 1. Consequence-preserving and consequence-changing variations in object-centric manipulation.** (a) Two global poses of the same fruit–blade interaction produce equivalent material-removal consequences in fruit-attached coordinates. (b) Changing the local peel thickness under an identical blade motion and removal-depth field changes the relationship between residual peel and exposed flesh. The two cases illustrate the complementary requirements of invariance to consequence-irrelevant spatial transformations and retention of physical information necessary for consequence prediction.

## 2.3 Geometric peeling simulator and outcome measures

A deterministic geometric simulator generates ground-truth peeling consequences. The fruit is represented by a triangulated, smoothly deformed sphere. Each direction $\mathbf{u}$ is a unit vector from the fruit center, and $p(\mathbf{u})$ is the corresponding point on its surface. The radius is varied along different directions according to

$$p(\mathbf{u}) = r(\mathbf{u})\mathbf{u}, \qquad r(\mathbf{u}) = r_0[1 + \textstyle\sum_{k=1}^{10} c_k\, \phi_k(\mathbf{u})]. \tag{12}$$

Here $r(\mathbf{u})$ is the distance from the fruit center to the surface, and $r_0$ is the nominal fruit radius of 40 mm. The ten functions $\phi_k(\mathbf{u})$ are prescribed, dimensionless smooth directional polynomial modes, and $c_k$ are independently sampled, dimensionless shape coefficients. The full definitions and coefficient scales are provided in Appendix A. The generator enforces a minimum radial distance of $0.82r_0$. Main learning experiments use 642 vertices and 1280 faces. The geometric invariance check uses 2562 vertices and 5120 faces.

The peel-thickness field is generated independently of the fruit shape using another set of smooth directional modes

$$h(\mathbf{u}) = \mathrm{clip}\big(h_0\big[1 + \textstyle\sum_{j=1}^{8} b_j\, \chi_j(\mathbf{u})\big], 0.60h_0, 1.45h_0\big). \tag{13}$$

Here $h_0$ is the nominal peel thickness of 1.5 mm. The eight functions $\chi_j(u)$ are prescribed, dimensionless thickness-variation modes, and $b_j$ are independently sampled, dimensionless coefficients. The exact modes and coefficient scales appear in Appendix A. The function clip($x$, $a$, $b$) restricts $x$ to the interval [$a$, $b$]. The thickness therefore lies between $0.60h_0$ and $1.45h_0$. A surface point s corresponds to $p(\mathbf{u})$, so $h(s)$ and $h(\mathbf{u})$ denote the same physical thickness field in two equivalent parameterizations. The symbols $b_j$ and $\chi_j$ are distinct from blade penetration d and stroke direction $\psi$.

Each stroke follows a surface path of prescribed length $L$ with the blade at a specified penetration $d$ and inclination $\alpha$. In the geometric oracle, inclination changes the penetration gradually along the stroke. The oracle computes removal from the three-dimensional blade trajectory and finite-width footprint. The resulting consequences are evaluated before the same physical interaction is encoded in alternative coordinate frames. This ordering keeps the ground-truth calculation independent of the representation. **Figure 2** illustrates the simulated fruit–blade interaction, the local cutting configuration, and the resulting spatial pattern of material removal. The consequence field is computed by the geometric oracle before the interaction is encoded in alternative coordinate representations.

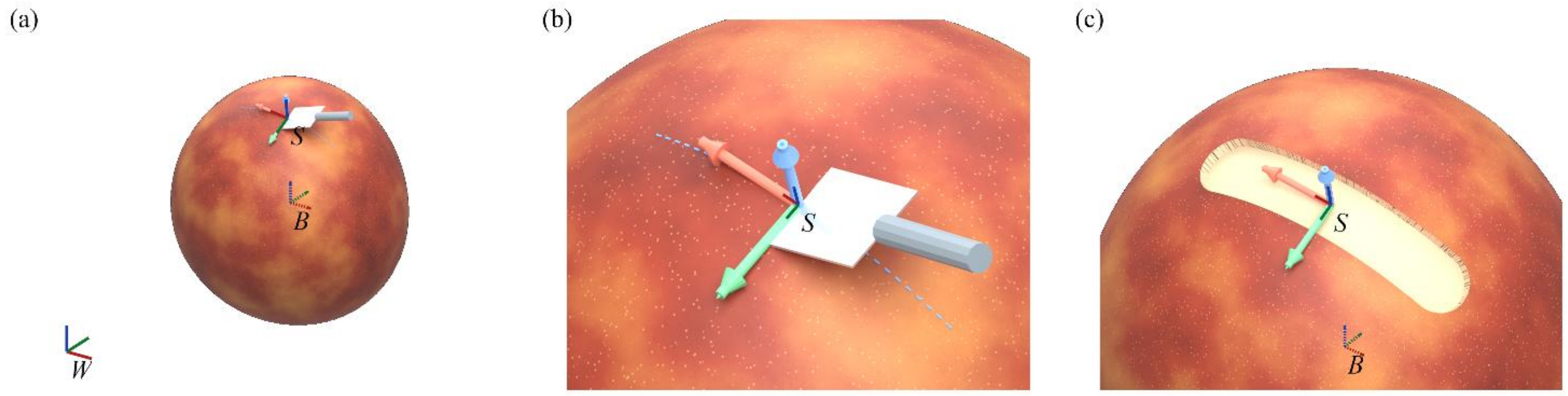

**Figure 2. Procedural peeling simulation and spatially resolved material removal.** (a) An irregular fruit surface and finite-width cutting blade represented in world, fruit-body, and local-surface coordinates. (b) A magnified view of the blade–surface interaction and local cutting configuration. (c) The resulting material-removal band following the prescribed surface stroke. The geometric oracle evaluates the cutting consequence independently of the coordinate representation supplied to the predictive model.

The simulator was checked by applying 100 shared rigid transformations to the complete fruit–blade interaction. The maximum change in the dense removal field was $6.38 \times 10^{-17}$ m. This numerical test verifies the required invariance of the generated labels to common global transforms.

Peeling quality is evaluated over the interior surface region fully swept by the blade during the prescribed action. This evaluation region is denoted by $\Omega_{\mathrm{eval}}$, with $|\Omega_{\mathrm{eval}}|$ representing its surface area. The differential surface-area element is denoted by d$S$. Coverage is defined as the fraction of this region where the residual peel thickness does not exceed a prescribed fraction $\varepsilon = 0.10$ of the original local thickness.

$$C_\epsilon = \frac{1}{|\Omega_{\text{eval}}|}\int_{\Omega_{\text{eval}}} \mathbf{1}\,[h(s) - q(s) \leq \epsilon h(s)]\mathrm{d}S. \tag{14}$$

Residual peel $U$ and flesh loss $O$ are the area-averaged depths of material left above or removed beyond the peel–flesh boundary

$$U = \frac{1}{|\Omega_{\text{eval}}|}\int_{\Omega_{\text{eval}}} \max\,(h(s) - q(s), 0)\mathrm{d}S, \tag{15}$$

$$O = \frac{1}{|\Omega_{\text{eval}}|}\int_{\Omega_{\text{eval}}} \max\,(q(s) - h(s), 0)\mathrm{d}S. \tag{16}$$

Both $U$ and $O$ have units of length and are reported in millimeters. Their separate definitions distinguish insufficient penetration from excessive material removal.

The relative depth error is $e(s) = q(s)/h(s) - 1$. Its area-weighted mean, denoted by $\bar{e}$, is the surface integral of $e(s)$ over $\Omega_{\text{eval}}$ divided by $|\Omega_{\text{eval}}|$. The depth-uniformity measure $V$ is the area-weighted standard deviation of this relative error

$$V = \sqrt{\frac{1}{|\Omega_{\text{eval}}|}\int_{\Omega_{\text{eval}}} (e(s) - \bar{e})^2\,\mathrm{d}S}. \tag{17}$$

The quantity $V$ is dimensionless. The outcome vector uses the shorthand $C = C_\varepsilon$ for the fixed tolerance $\varepsilon = 0.10$. The four reported targets are coverage $C$, residual peel $U$, flesh loss $O$, and depth nonuniformity $V$. All four outputs are reported separately. In particular, coverage alone can increase through excessive penetration. The local thickness field changes the quality of a given removal depth even when blade geometry is held fixed.

## 2.4 Predictive models and controlled evaluation

All representation comparisons use the same action-conditioned PointNet-style predictor and exactly the same underlying physical samples. Each sampled surface point carries seven features comprising three position coordinates, three normal components, and peel thickness. Eight sampled positions along the blade path, together with blade width and blade angle, form the action input. The action encoder has hidden widths 64 and 48. Its output is concatenated to every point. Pointwise layers of widths 96, 96, and 64 are followed by concatenated max and mean pooling, and an output head of widths 96 and 64 predicts $[C,U,O,V]$. The architecture is intentionally modest because network capacity is not the contribution.

The same selected surface points are used across world, object, and surface-relative inputs. Positions, normals, and blade-path coordinates are transformed and peel thickness is unchanged. Inputs use representation-independent physical scaling rather than per-representation z-scoring. This choice is important for the frame-noise experiment because a coordinate that is nearly constant after canonicalization can have an arbitrarily small empirical standard deviation. Dividing by that value would amplify small perturbations for purely numerical reasons. Output channels are standardized from the training split. For reporting, aggregate NRMSE is the root mean of squared component errors after division by the test-set standard deviation of each consequence channel. Component-wise RMSE values are also retained so that the aggregate score does not hide which physical quantity drives a difference.

Models are optimized with AdamW and mean-squared error under the same learning rate, weight decay, minibatch size, model-selection criterion, and physical train/validation/test split within each comparison. The primary baselines are world-, object-, and surface-relative coordinates. A world-frame rigid-transform augmentation control retains the same physical labels but adds four coordinate views per training interaction. This baseline tests whether a straightforward finite augmentation can substitute for encoding a known rigid relation analytically. Architecture-specific equivariant policies are not used as a performance baseline because they would change both the representation and the hypothesis class. The objective here is to isolate the effect of parameterizing the same physical interaction differently, not to claim state-of-the-art predictive accuracy. Exact hyperparameters and experiment-specific patch sizes are reported in Appendix A.

The controlled tests vary global pose, equivalent surface-contact location, peel thickness, local-frame estimation and the physical motion of the blade. Frame-origin uncertainty is implemented by subtracting the same random displacement from surface-point positions and sampled blade positions after defining the true action. The Oracle label, actual tool path, and patch identity remain fixed. Translation noise $\sigma$ denotes the Gaussian standard deviation of each active local coordinate component, not a fixed Euclidean displacement (the three-axis RMS magnitude is $\sqrt{3}\sigma$). Frame rotations are applied jointly to local point positions, normals and blade positions. Further controls displace the estimated contact location, optionally reselecting its local surface patch, or change actual blade penetration and recompute the Oracle outcome. These distinct interventions separate coordinate instability from missing perception information and changed physical consequences.

The global-pose and equivalent-contact tests use a sphere with constant thickness, with 600 training, 150 validation and 300 test interactions per experiment. The irregular-fruit frame study uses 14 fruits and 700 training, 140 validation and 280 test interactions. The independent mechanistic robustness analysis regenerates three fruit-bank cohorts with these irregular-fruit sample counts and five model initializations per cohort. Each frozen predictor is tested with repeated random perturbations; test-noise repeats are averaged within each trained model, rather than treated as additional independent training seeds. The physical-sufficiency comparison uses 14 base geometries, 1000 training interactions, 220 validation interactions and 90 paired thickness interventions. The decision evaluation uses 72 shared scenes with 18 candidate strokes per scene. The lower thickness-scaling factor in this paired intervention is $0.8h$ relative to the nominal thickness. The upper factor is 1.2.

The decision experiment evaluates a shared finite set of candidate strokes, denoted by $\mathcal{A}$, from the same initial state. The simulator provides the true cost $J(a)$ for each action $a$ in $\mathcal{A}$. The learned model computes an estimated cost $\hat{J}(a)$ from its predicted peeling outcomes. The ground-truth and model-selected actions are

$$a_{\mathrm{GT}} = \underset{a\in\mathcal{A}}{\operatorname{argmin}} J(a), \qquad a_{\mathrm{WM}} = \underset{a\in\mathcal{A}}{\operatorname{argmin}} \hat{J}(a), \tag{18}$$

Decision regret $\Delta J$ measures the extra true cost incurred by choosing the model-preferred action rather than the oracle-preferred action

$$\Delta J = J(a_{\mathrm{WM}}) - J(a_{\mathrm{GT}}). \tag{19}$$

For decision evaluation, 72 shared irregular-fruit scenes are generated. Each contains 18 candidate actions formed by six penetration depths and three inclination angles. Other scene and stroke variables are held fixed. The cost used for both oracle and model ranking is

$$J = \frac{U}{h_{\mathrm{ref}}} + 2\frac{O}{h_{\mathrm{ref}}} + 0.35V, \qquad h_{\mathrm{ref}} = 1.5\ \mathrm{mm}, \tag{20}$$

The reference thickness $h_{\mathrm{ref}}$ converts the mean residual-peel and flesh-loss depths into dimensionless quantities. The coefficients 1, 2, and 0.35 are fixed for all representations and assign greater penalty to flesh loss than residual peel. Predicted cost $\hat{J}$ uses the same expression with model-predicted $U$, $O$, and $V$.

Decision quality is evaluated by pairwise ordering accuracy, Spearman rank correlation, oracle-best-in-top-3 recall, and normalized regret. Pairwise comparisons omit near-ties with true cost differences at most 5% of the cost range within a scene. Normalized regret divides the nonnegative cost difference $\Delta J$ by the difference between the largest and smallest true candidate costs in that scene. Scenes with a zero cost range are omitted from this ratio. Exact top-1 recovery is retained as a diagnostic because several candidate sets contain near-optimal neighboring actions.
Principal representation comparisons reuse the same training-initialization seeds and are reported as paired five-seed contrasts, including effect size and its direction across seeds. With five nonzero paired differences, the smallest attainable two-sided exact Wilcoxon $p$ value is 0.0625. Statistical interpretation therefore rests primarily on the paired effects and response curves. The robustness analysis adds three independently regenerated fruit/sample cohorts, each with five fitted model seeds. Repeated perturbation draws on a single frozen model are not independent fitted models, and the three cohorts remain the limited units of physical-generation replication. Noise-response crossovers are descriptive interpolations between tested error levels rather than physical critical tolerances. The equal-update-budget augmentation comparison uses two fruit cohorts with the number of optimizer updates matched within each original/augmented seed pair.

# 3. Results

## 3.1 Finite-data cost of consequence-preserving spatial variation

The rigid-pose experiment tests the first requirement at its most transparent limit. A blade follows the same path relative to a fruit while the complete fruit–blade system changes its world position and orientation. No new physical cutting relation has been introduced. The geometric oracle returns the same material-removal field in fruit-attached coordinates. Across 100 independent shared rigid transformations the maximum change is $6.38\times10^{-17}$ m. This establishes physical equivalence independently of the learned predictor. Any subsequent change in predictive error must therefore arise from the coordinate description or the learner's treatment of it, rather than from altered material removal.
The World and Object baselines coincide when global pose variation is absent, both obtaining a five-seed mean NRMSE of 0.8584 (**Figure 3** (a)). This zero-range agreement provides a control for differences in the subsequent pose sweep. Because the Oracle consequences and physical interactions are matched, a divergence induced by increasing pose variation identifies a learning burden associated with input coordinates rather than a change in material removal.

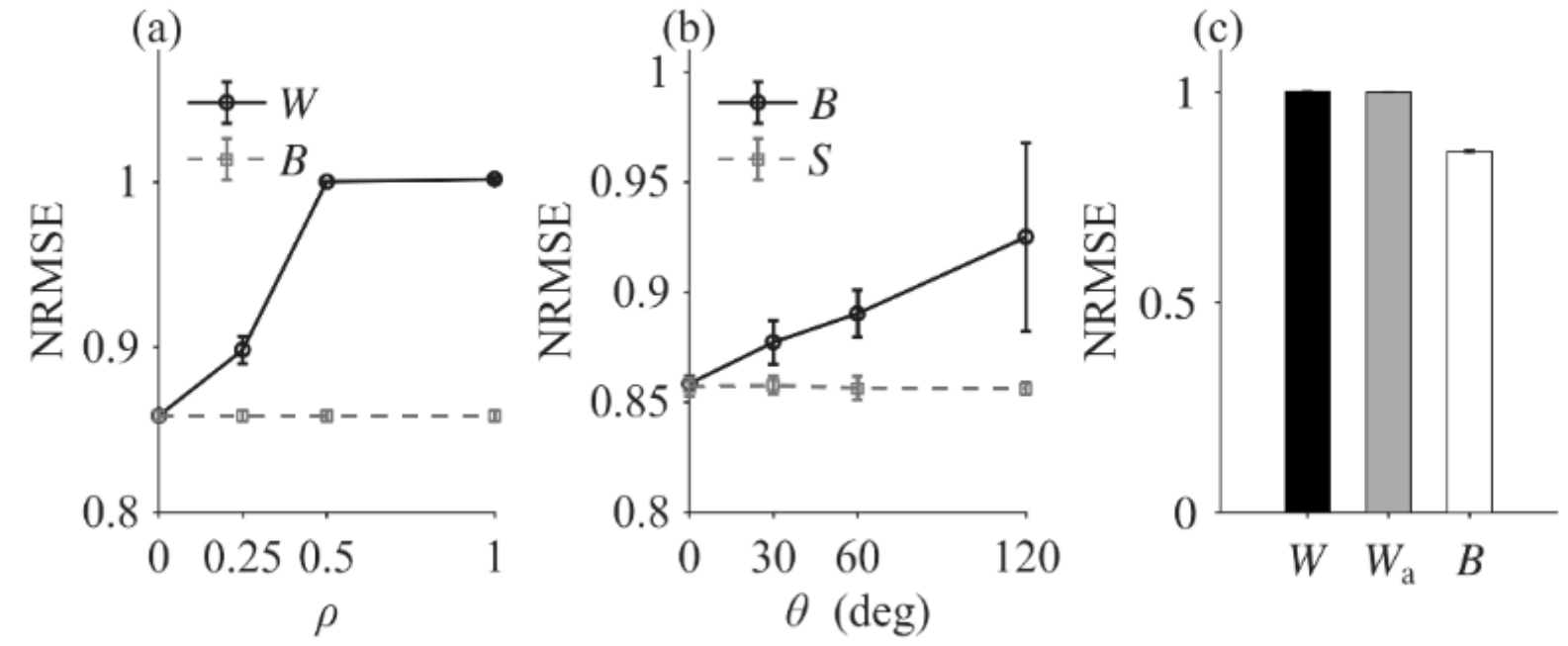

**Figure 3. Prediction accuracy under physically equivalent pose and contact variation depends on the chosen spatial action representation and the available training data.** (a) World-frame (*W*) and body-frame (*B*) prediction error with increasing common global pose range $\rho$. (b) Body-frame (*B*) and surface-frame (*S*) error under equivalent contact relocation through angle $\theta$. (c) Predictions from *W*, world coordinates with four augmented views per training sample ($W_a$), and *B*. Points and bars indicate five-seed means; error bars show standard deviations across independently trained seeds.

Increasing pose range separates the representations. World NRMSE increases from 0.8584 to 0.8984 at $\rho$ = 0.25, reaches 1.0004 at $\rho$ = 0.50 and 1.0017 at the largest range. Object remains at 0.8584 because the common motion is canceled by construction. The largest paired World–Object difference is 0.1433 NRMSE, or approximately 16.7% of Object error. All five seeds favor Object (Table B.1). The World curve approaches a plateau, so the experiment identifies a finite-range learning penalty rather than an unrestricted scaling law.

Finite exposure to transformed examples only partly addresses this redundancy. The World + augmentation control creates four coordinate views from the same 600 physical training interactions. Its five-seed NRMSE is 1.0002, close to the unaugmented World value of 1.0017, while Object attains 0.8584 (Figure 3(c)). Because additional views also change the number of training updates, this is a same-physical-data comparison rather than a matched-budget test. It establishes neither the inferiority of general augmentation nor the behavior of specialized equivariant architectures. It demonstrates the benefit of encoding this known global rigid relation directly in the tested finite-data learner.

The next experiment tests spatial redundancy within the object frame. On a symmetric fruit with uniform thickness, the same local blade interaction is transferred between surface positions. Global fruit pose is held fixed. Under the idealized symmetry, each relocation preserves the local geometry and prescribed cutting relation. Nevertheless, Object-frame contact positions and blade orientations change. Surface-relative coordinates instead express the action with respect to the contacted surface. This paired construction isolates the additional coordinate variability that remains after global pose normalization.

Without contact relocation, Object and Surface obtain similar NRMSE values of 0.8584 and 0.8572 (Figure 3(b)). With increasing relocation through 30°, 60° and 120°, Object error becomes 0.8772, 0.8904 and 0.9250. Surface remains near 0.857 throughout the sweep, ending at 0.8561. At 120°, Surface improves NRMSE by 0.0689, approximately 7.4% relative to Object, with the same effect direction in all five seeds (Table B.1). The separation demonstrates a second coordinate-related learning burden that persists after global object-pose normalization.

The two mechanism tests operate at different spatial scales and should be interpreted together. The World–Object comparison removes the global placement of an interacting body. The Object–Surface comparison removes the location of an equivalent interaction on a symmetric object. Neither experiment changes the physical information available to its matched predictors. These results support the value of expressing actions in coordinates aligned with the invariance of the target consequence. Surface relocation on irregular fruits is only approximately equivalent because local curvature and thickness can differ. The following experiment tests the complementary requirement that physically relevant differences remain available to the model.

## 3.2 Preserving consequence-relevant physical information

A useful representation must retain variations that affect the predicted consequence. Peel thickness offers a controlled intervention for testing this requirement. Two versions of the same fruit receive identical blade trajectories at identical contacts and global poses. Their thickness fields differ by scale factors of 0.8 and 1.2. The geometric blade sweep is consequently unchanged, but the amount of remaining peel and the amount of underlying material exposed by that sweep differ. This experiment separates a physical quality change from a coordinate change without introducing a second action or a more complex cutting model.

The two predictors have the same architecture, physical action inputs and input dimensions. The complete Surface model receives the local thickness field. The thickness-blind Surface ablation replaces that channel with a constant. For every paired intervention, the blind predictor receives two numerically identical inputs even though the simulator assigns different peeling-quality outcomes. This exact input collision is the central result of the test. It establishes a failure of predictive sufficiency independently of the learning algorithm and shows why maximal coordinate invariance is not a suitable objective.

The paired physical effect remains substantial under the unified experiment series. Across 90 otherwise identical blade sweeps, increasing peel-thickness scale from 0.8 to 1.2 changes mean coverage by −0.247, increases residual peel by 0.271 mm, decreases flesh loss by 0.336 mm and changes nonuniformity by −0.128. Thus a thicker peel layer leaves more material above the target boundary and reduces the amount counted as overcutting. These differences arise from a material-state intervention, not a change in the blade path or global pose.

The thickness-blind predictor necessarily assigns identical outputs to each paired intervention because its two input tensors collide exactly. Retaining thickness removes that constructed ambiguity; the resulting finite-network predictions need not perfectly recover the full componentwise response. These paired physical-response statistics concern 90 intervention pairs and should not be confused with the separate five-seed prediction-error contrast.

Across five training seeds, full Surface prediction attains an NRMSE of 0.3888 ± 0.0100, compared with 0.8596 ± 0.0488 when thickness is withheld. The paired error difference is 0.4708 and all five seeds favor the complete input (Figure 4(a), Table B.1). This roughly 54.8% reduction relative to the blind model follows from the unified experimental protocol. The result complements rather than replaces the more fundamental input-collision argument: better coordinate normalization cannot restore a physical variable deliberately omitted from the input.

The collision imposes a model-independent limit. Any deterministic predictor assigns a common prediction to two identical represented inputs. For two equally weighted outcomes expressed on the same fixed dimensionless scale, their average squared Euclidean error is at least one quarter of the squared separation between the true outcomes (Eq. 11). This is a per-pair inequality, not a universal error floor. It identifies missing consequence-relevant information as a failure that neither more optimization nor a higher-capacity predictor can eliminate without changing the representation.

Transfer to different procedural fruits gives a useful boundary on geometric locality. For the Surface model, the three-seed mean standardized prediction error is 0.4367 for new strokes on seen fruit identities and 0.3860 on unseen fruits whose local geometry remains familiar, but increases to 3.8493 for unfamiliar curvature and size. The corresponding Object errors are 0.8757, 0.8409 and 4.1642. The large penalty under geometric distribution shift indicates that a useful local coordinate frame does not guarantee extrapolation to previously unrepresented contact geometry. This supporting three-seed experiment is distinct from the core five-seed contrasts and is not used to claim a universal scaling law.

The thickness and shape tests show where the benefits of coordinate simplification must stop. Surface-relative actions can suppress the placement of a contact while retaining the thickness and geometry of that contact. The thickness-blind failure does not implicate Surface coordinates themselves. It demonstrates the distinct cost of withholding a variable that determines the evaluation outcome. This distinction is essential to the central discovery, because lower coordinate variation is valuable only when the retained representation remains informative about consequence.

## 3.3 Directional frame uncertainty and the limits of canonicalization

Accurate surface frames remove a large part of the irrelevant spatial variation in peeling. Their estimation errors, however, need not represent actual errors in the physical interaction. The unified frame-boundary experiment perturbs the estimated local origin while leaving the sampled physical patch, blade sweep and removal consequences unchanged. If a local surface point $p$ and blade-path point $a$ are both represented relative to an erroneous origin $o + \delta$, their local coordinates change by the same displacement, while $(a - \delta) - (p - \delta) = a - p$. Thus the physical blade–surface separation is preserved exactly. A generic PointNet operating on absolute local coordinates is not guaranteed to respect this common-translation symmetry. The experiment therefore tests a potential failure of the learned representation rather than a mechanical failure of peeling.

Under the accurate frame of the unified five-seed irregular-fruit experiment, Surface reaches NRMSE 0.5678 ± 0.1312, compared with 0.8951 ± 0.0106 for Object. The clean Surface advantage corresponds to an approximately 36.6% reduction relative to Object. With independent Gaussian origin perturbations applied to each of the three local coordinates, Surface NRMSE rises to 1.0324 at $\sigma = 1$ mm and 1.4583 at $\sigma = 2$ mm (**Figure 4** (b)). The average prediction advantage therefore reverses between 0 and 1 mm in this specified protocol. Here $\sigma$ is a per-axis standard deviation: a nominal 1 mm three-axis error has a Euclidean RMS magnitude of approximately 1.73 mm. These values characterize model sensitivity under a defined stochastic coordinate perturbation, not the blade's positioning tolerance. The response continues to diverge from the fixed clean Object benchmark as $\sigma$ increases: Surface NRMSE is 2.2914 at 6 mm, 2.7989 at 8 mm and 3.4659 at 12 mm. The apparent onset of this disadvantage is not a fixed property of the geometry or an intrinsic material threshold. The observed response is a joint property of the perturbation law, its direction, the trained predictor and the choice of clean comparator. The independent directional and invariance tests below separate these factors and show why an error interval obtained in one protocol should not be generalized to another.

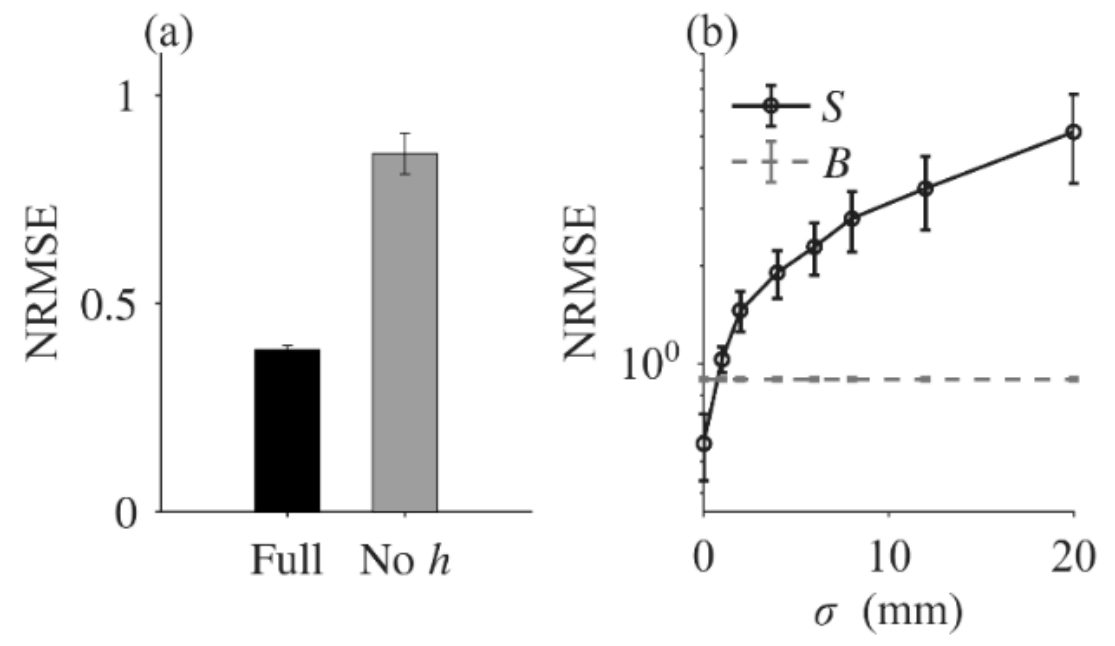

**Figure 4. Prediction error depends on retaining physically relevant local state and on the accuracy of the estimated surface-frame origin.** (a) Prediction NRMSE when local peel thickness $h$ is retained (Full) or omitted (No $h$). (b) Surface-relative ($S$) and body-relative ($B$) prediction NRMSE under isotropic Gaussian perturbations of the estimated local-frame origin. Results are five-seed means with standard-deviation error bars. Panel (b) uses a logarithmic NRMSE axis.

To distinguish statistical noise from a directional mechanism, three additional, independently regenerated fruit-bank cohorts were evaluated using five model seeds per cohort and three test-noise draws for each frozen model. The median first Surface–Object crossing occurs at $\sigma$ = 0.840 mm for isotropic origin noise and 0.829 mm for normal-only noise, compared with 6.829 mm for tangential-only noise among the 13 of 15 models that cross within the 12 mm range. Under isotropic noise, 14 of 15 models cross before 1 mm. The remaining fitted model crosses near 6.21 mm but begins from much worse clean-frame accuracy. The 6–8 mm interval can therefore arise under tangentially dominated errors, yet a later crossing does not by itself identify a more accurate model. Normal-origin perturbations are more harmful than tangential ones in all 15 tested predictors at matched per-component σ, a response that cannot be explained by random initialization alone.

A direct invariance control sharpens this result. Subtracting the sampled surface-patch centroid from every patch point and the blade path removes their common origin translation algebraically. Across all 15 independently fitted centered models, the mean NRMSE varies by less than $5 \times 10^{-8}$ across the tested 0 -12 mm translation levels, and none loses its advantage over paired clean Object. Centering therefore eliminates this specific representational nuisance, but it is not a universal remedy. It leaves unchanged the need to estimate surface orientation, identify the correct patch and observe the executed action. The controlled mechanisms in **Figure 5** separate these limits.

Orientation errors exhibit another directional asymmetry. With frozen pretrained models, perturbing the local frame by rotating about its normal leaves Surface NRMSE near 0.516 up to 15°, whereas tilting the estimated normal increases it to 0.759 at 6°, 0.949 at 10° and 1.140 at 15° (**Figure 5**(a)). Centered Surface is less sensitive to this tilt but is not rotation-invariant. When contact mislocalization instead reselects local surface points, error rises despite identical true blade motion. Surface reaches 1.011 and the centered variant 1.185 at 6 mm tangential-location noise (**Figure 5**(b)). These tests show why invariance to a common coordinate shift must be distinguished from sufficiency of the observed geometry. A representation cannot recover a missing physical patch merely by canceling its coordinate origin.

Recomputing the physical Oracle after perturbing executed blade penetration distinguishes action deviations from representational noise. At 0.5 mm execution-error standard deviation, normalized true consequence change is 0.811.

A Surface predictor given only the commanded trajectory has NRMSE 0.916, compared with 0.636 when the measured executed trajectory is supplied (**Figure 5** (c)). The remaining difference reflects finite-model prediction error. Small lateral execution changes have a weaker average effect under the tested action distribution. The comparison establishes that physical consequence invariance cannot be demanded when the executed intervention itself changes.

A matched-optimizer-update augmentation control further demonstrates that robustness and clean-frame accuracy are competing objectives rather than a single scalar benefit. Across two independently generated fruit/sample cohorts, the unaugmented Surface predictor changes from 0.559 clean NRMSE to 2.446 at 6 mm isotropic origin noise. Augmenting with a 2 mm origin-noise distribution changes these values to 0.833 and 0.924, respectively (**Figure 5** (d)). Since the number of physical samples and optimizer updates is paired within each seed, this shift cannot be attributed solely to an enlarged update budget. It is a change in the learned input invariance, paid for with lower clean precision.

Across the numerical tests, a causal decomposition emerges. An unchanged physical relation expressed in unstable coordinates calls for an invariant encoding. An incorrectly selected patch calls for better physical observation. And an actual deviation from commanded tool motion calls for knowledge of the executed action. These failure modes may give superficially similar prediction-error curves, but their correct remedies differ. In particular, a single location-error threshold conflates the direction of frame uncertainty, the completeness of perceived local geometry and the predictive invariance learned during training.

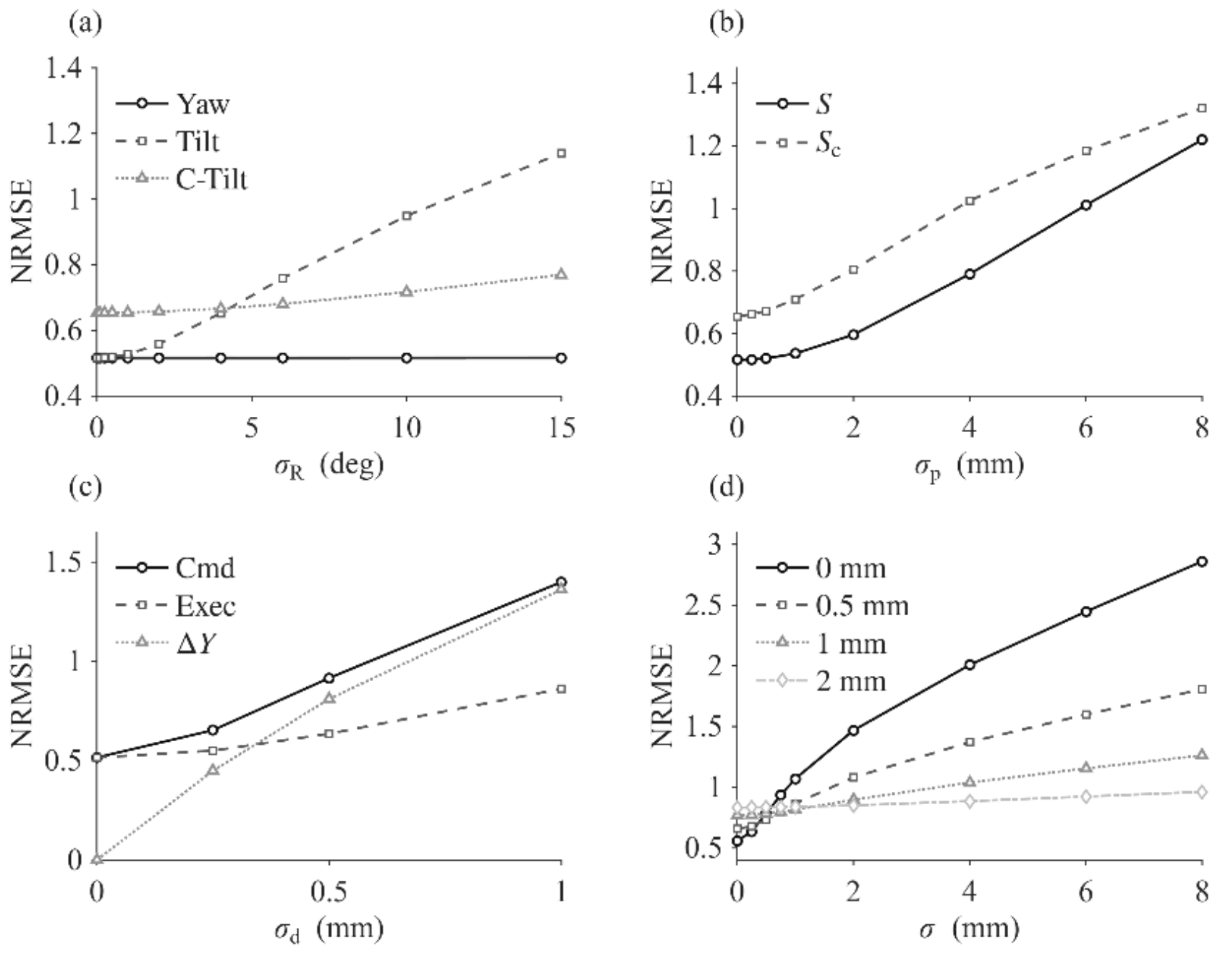

**Figure 5. Surface-relative prediction responds differently to orientation errors, contact mislocalization, execution deviations, and noise augmentation.** (a) Orientation errors about the surface normal (Yaw), normal-direction tilt (Tilt), and tilt with joint surface-patch/tool-centroid recentering (C-Tilt). (b) Contact-patch mislocalization for the surface-relative predictor ($S$) and its centered form ($S_c$). (c) Predictions using the commanded stroke (Cmd) or executed stroke (Exec), alongside the normalized change in geometric-oracle outcome ($\Delta Y$). (d) Test-noise response for models trained with different coordinate-noise augmentation levels under an equal update budget. Panels (a)-(c) use three fruit cohorts with five independently trained models per cohort. Panel (d) uses two cohorts with five models each. Curves show averages over fitted models. Repeated noise draws are averaged within models.

## 3.4 Consequences for counterfactual action selection

A robot world model should support the comparison of alternative actions before execution. This role motivates a decision-level evaluation beyond the average error of its predicted quality variables. The test uses 72 irregular-fruit scenes. Each scene provides 18 counterfactual blade strokes formed from six penetration levels and three inclination angles. The fruit, contact point, tangential direction, stroke length, blade width and global pose are held fixed within a scene. The simulator evaluates all candidate actions from the same initial state. This design provides an oracle ordering without confounding candidate selection with grasp planning or long-horizon control.

The candidate set probes a physically meaningful decision boundary. A shallow stroke leaves peel behind, while a deep stroke can remove underlying material. The selected blade inclination further changes the spatial pattern of removal. The comparison therefore requires the predictor to distinguish modest changes in physical action variables rather than merely recognize different global poses. The cost uses fixed weights on residual peel, flesh loss and depth uniformity. Its purpose is to test whether predicted consequences retain a usable ordering among the same candidates for every representation.

Under accurate local frames, the three representations exhibit different ranking fidelity in the unified five-seed decision experiment (Table 1 and **Figure 6** ). Pairwise ordering accuracy is 0.484 ± 0.018 for World, 0.681 ± 0.024 for Object and 0.729 ± 0.089 for Surface. Surface improves broad action ordering over World, while Object improves pairwise ordering without necessarily reducing the cost of the finally selected action. The results demonstrate that a representation should be judged by the decision criterion it serves, not only its mean regression error.

Table 1. Counterfactual prediction and action-selection performance of the three representations under accurate coordinate frames in the unified five-seed experiment. Values are mean ± standard deviation across independent model seeds. Higher pairwise accuracy, Spearman correlation and top-three recall are preferable. Lower prediction NRMSE and normalized regret are preferable. The different criteria need not induce the same representation ordering.

| Representation | Prediction NRMSE ↓ | Pairwise ranking ↑ | Spearman ↑ | Oracle in Top-3 ↑ | Normalized regret ↓ |
|---|---|---|---|---|---|
| World | 1.025 ± 0.001 | 0.484 ± 0.018 | −0.033 ± 0.046 | 0.158 ± 0.065 | 0.653 ± 0.038 |
| Object | 0.898 ± 0.012 | 0.681 ± 0.024 | 0.434 ± 0.036 | 0.503 ± 0.139 | 0.681 ± 0.021 |
| Surface-relative | 0.561 ± 0.109 | 0.729 ± 0.089 | 0.526 ± 0.160 | 0.772 ± 0.258 | 0.226 ± 0.176 |

Rank correlation is −0.033 for World, 0.434 for Object and 0.526 for Surface. Oracle-best-in-top-three recall is 15.8%, 50.3% and 77.2%, respectively. These distinct measures distinguish identifying broadly preferable actions from selecting a small subset containing the oracle optimum. Object's intermediate ranking fidelity does not imply intermediate regret: a predictor may rank many pairs correctly while still choosing a costly action when its errors concentrate near the optimum.

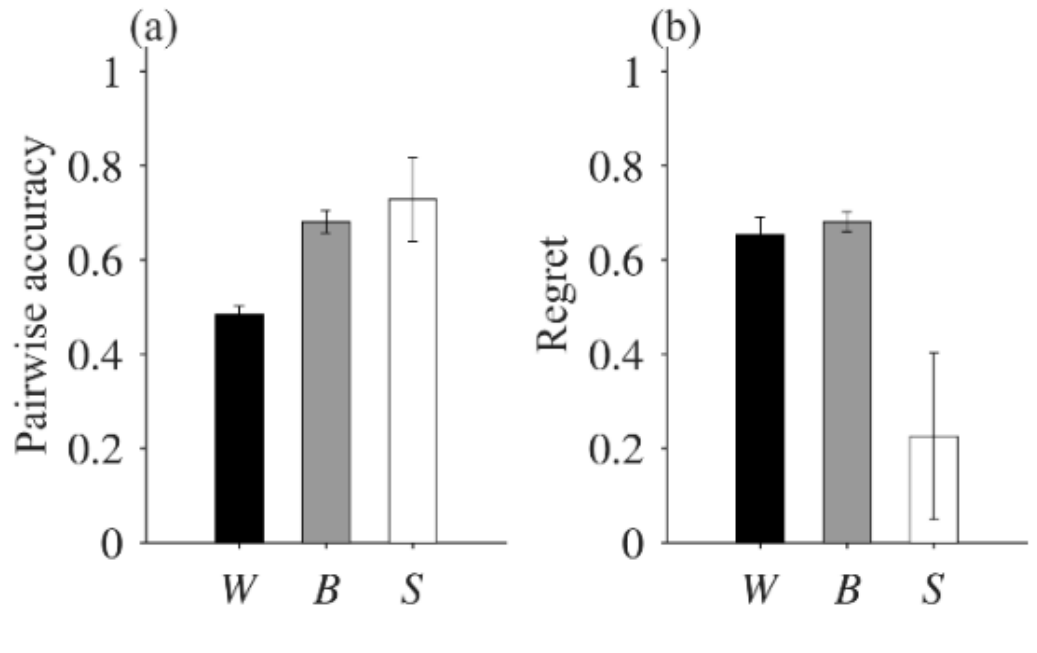

**Figure 6. World-, body-, and surface-relative representations yield different counterfactual action-ranking accuracy and decision regret.** (a) Pairwise action-ordering accuracy and (b) normalized decision regret for world (*W*), fruit-body (*B*), and surface-relative (*S*) predictors under accurate local frames. Bars show means across five independent training seeds. Error bars indicate between-seed standard deviations.

Surface achieves mean normalized regret 0.226, compared with 0.653 for World and 0.681 for Object. Relative to World, Surface lowers regret by approximately 65.3%. Object has slightly higher mean regret than World despite higher pairwise accuracy, underlining the distinction between global ordering and the consequence of the selected candidate. Regret is computed from shared true action costs and the same scene-wise normalization. The comparison supports a decision-level benefit of physically aligned local inputs in this candidate set without establishing universal dominance in robotic planning.

Exact oracle-best recovery remains an informative but distinct criterion. In the updated five-seed evaluation, mean top-one recovery is 5.0% for World, 0.6% for Object and 37.8% for Surface. The dispersion of Surface performance is substantial across seeds, and neighboring low-cost actions may have similar consequences. Pairwise fidelity, oracle-in-top-three recall and normalized regret therefore provide complementary evidence. None should be replaced by an unqualified top-one success claim.

Prediction and action ranking respond differently to frame error, but the direction of the decision-level effect must be established from the same experimental batch. In the unified decision-noise sweep, Surface pairwise accuracy declines from 0.729 under clean frames to 0.665 at 2 mm and 0.555 at 12 mm. Normalized regret increases from 0.226 to 0.412 and 0.568. At 12 mm, Surface accuracy falls below the clean Object baseline of 0.681, although Surface mean regret remains below the Object value of 0.681 in this particular evaluation. Consequently, the old assertion that Surface retains a ranking advantage throughout the noise sweep is not supported by the unified data. Neither the prediction-error crossover nor a ranking crossover supplies a universal decision threshold across different datasets.

The decision-theoretic mechanism depends on where prediction errors occur within the candidate cost landscape. A model can have biased absolute outcome estimates and still correctly order many alternatives. Conversely, modest average error can reverse the selection between near-optimal candidates. The contrast between Object's ranking and regret illustrates this asymmetry. Frame uncertainty further shows that a benefit demonstrated under accurate coordinates cannot be extrapolated automatically to error-corrupted counterfactual comparisons. This strengthens the case for reporting predicted physical consequences together with action-level metrics, with a clear statement of which frame errors affect only the input and which alter the executed action itself.

## 4. Discussion

The controlled comparisons show that spatial generalization in action-conditioned world models depends on more than the choice of a global or local reference frame. A representation may retain coordinate variation that leaves the physical interaction unchanged, making shared relationships harder to learn from finite data. It may instead discard physical differences that determine the outcome, leaving the predictor unable to distinguish correct answers. Even a physically appropriate representation can fail when the learned model remains sensitive to irrelevant coordinate changes. These findings distinguish three limitations of spatial generalization and explain why geometric normalization alone is neither a complete solution nor an appropriate objective in itself.

When a fruit and blade undergo a common rigid transformation, their fruit-attached cutting relationship is unchanged, although the world-coordinate description varies. Object coordinates remove this variation directly, and Surface coordinates additionally align equivalent contacts on a symmetric object. The resulting performance differences indicate that representing a known physical equivalence can reduce the learning burden imposed by finite spatial coverage. They do not establish that World predictors are fundamentally incapable of generalization. Sufficiently broad training data, targeted augmentation, or equivariant architectures may learn the same relation. The advantage of suitable relative coordinates is that they expose part of this structure before learning rather than requiring every such relation to be inferred from examples.

This result complements advances in world models that learn dynamics through imagined trajectories and structured observations [1-3, 5]. Visual and three-dimensional approaches have further improved the representation of scene evolution[6, 8, 9], while object-centric models organize predictions around interacting entities [11, 12]. The present experiments isolate a different aspect of the same predictive problem. Richer descriptions of an environment do not by themselves specify which interventions at different locations should share learned consequences. How actions are expressed relative to the interacting body or contact surface remains a separate modeling decision.

The findings also refine the interpretation of equivariant representations and canonicalization in manipulation. Structured object-relative descriptors and equivariant policies have demonstrated the benefits of incorporating geometric transformations into learning[15, 16, 18, 22, 23]. The relevant symmetry for consequence prediction, however, is defined by the physical process and the output of interest, not by geometry alone. Fruit-attached removal can remain invariant under a common transformation, whereas removal geometry expressed in world coordinates should transform equivariantly. Changes that alter gravity-relative loading or contact conditions need not preserve either relation. This task-dependent distinction is consistent with analyses of correct and incorrect equivariance [27]. Canonical coordinates are useful when they encode a justified physical equivalence rather than an assumed one.

The thickness experiment establishes the complementary requirement that an input must retain the information needed to predict an outcome. Identical blade paths over different peel thicknesses can produce different material removal. With thickness omitted from the complete predictive input, these interactions become indistinguishable, creating an irreducible error that additional training cannot remove. Predictive sufficiency therefore concerns the combined state and action information, not the action frame in isolation. The shape-transfer results reinforce this boundary. Surface-relative coordinates remain useful for unfamiliar fruit identities with familiar local geometry, yet prediction deteriorates for substantially shifted curvature and scale. A representation can reorganize available physical information, but cannot reconstruct a missing material state or guarantee extrapolation beyond the geometries encountered in training.

The frame-perturbation experiments reveal a different distinction between physical invariance and learned predictive behavior. Shifting the estimated coordinate origin changes both surface-point and blade coordinates without changing their relative geometry or the true removal outcome. Nevertheless, predictions from absolute local-coordinate inputs can deteriorate markedly. Centering the surface patch and blade path together cancels this shared translation algebraically and eliminates this specific sensitivity in the tested representation. Other errors behave differently. Perturbations along the surface normal are more damaging than tangential shifts, and normal tilt is more damaging than rotation about the normal. This asymmetry is consistent with the importance of local penetration geometry, although the measured sensitivity remains dependent on the trained model and perturbation distribution.

Coordinate changes must also be distinguished from errors in what the robot observes or executes. Selecting an incorrect contact patch replaces the geometry describing the actual interaction, which centering cannot restore. A deviation in executed blade penetration changes the physical consequence itself, so invariance to that deviation would be inappropriate. Predictions improve when the executed rather than merely commanded trajectory is supplied, because the input better reflects the intervention that occurred. These mechanisms call for different solutions, including appropriate coordinate invariance, reliable contact observation, and executed-action feedback. Their separation explains why a single numerical frame-error threshold cannot characterize the reliability of a surface-relative predictor.

The augmentation controls further show that robustness is not equivalent to accuracy under ideal observation. Finite global-pose augmentation produced limited improvement for the tested World predictor, whereas matched-update local-frame noise augmentation substantially improved tolerance to uncertain origins at the cost of clean-frame precision. The results reflect different ways of exposing a learner to equivalent descriptions. Analytic coordinate normalization removes a known nuisance directly, while augmentation alters the training distribution and the balance of errors across test conditions. Accordingly, the noise level at which Surface and Object performance crosses is a property of the predictor, augmentation procedure, and perturbation direction, not an intrinsic geometric tolerance.

These experiments do not imply that augmentation or equivariant architectures are generally inferior to relative coordinates.

The decision experiments connect these representation effects to the purpose of an action-conditioned world model. Surface-relative predictions improved candidate ordering and reduced selection regret under accurate local observations. Yet Object ordered more candidate pairs correctly than World while showing slightly higher mean regret. This apparent discrepancy is informative. Ranking accuracy averages many comparisons, whereas regret depends on the cost of the selected action and can be dominated by errors near favorable candidates. Frame uncertainty can likewise affect ranking and regret differently. Prediction error, candidate ordering, and realized selection cost therefore provide complementary rather than interchangeable measures of decision reliability.

These conclusions are bounded by the controlled simulator. Its geometric oracle models material removal under prescribed strokes, not fracture, adhesion, force transients, tool wear, or closed-loop control studied in broader peeling and cutting research [29, 30, 32, 33]. Contact relocation is exactly equivalent only under stipulated physical symmetries. The core comparisons use five paired model seeds, and extended perturbation tests use models nested within three independently generated fruit cohorts. Repeated noise draws are not independent physical replications. These experiments identify mechanisms, not universal effect sizes, hardware tolerances, or long-horizon robot performance.

The findings support a consequence-based approach rather than a preferred coordinate frame. A world model should identify transformations preserving its prediction target, retain physical state information that distinguishes different outcomes, and verify that learned predictions honor those relationships. Testing these requirements with measured surface geometry, tactile and force feedback, executed tool trajectories, and richer cutting mechanics is a natural next step, particularly for closed-loop and multistep tasks. More broadly, spatial generalization should be judged by whether learned action-state-consequence relationships transfer across configurations without confusing physical differences. This perspective connects the design of action representations to perception and action execution.

## 5. Conclusions

This work clarifies how action representation governs the transfer of learned physical relationships across spatial configurations in robot world models. The analysis identifies two distinct limitations of spatial generalization. Different coordinate descriptions of physically equivalent interactions increase the learning burden under limited training data, whereas representations that discard information essential to physical outcomes create prediction ambiguities that further training cannot resolve. The controlled comparisons show that object- and surface-relative coordinates facilitate the reuse of learned interaction relationships across changes in object pose and contact location, provided that the relevant physical state remains available. These results explain why spatial generalization cannot be achieved simply by increasing the complexity of scene representations or normalizing action coordinates.

A further finding is that physical equivalence does not necessarily translate into consistent predictions from a learned model. Changes in estimated local frames can alter predictions without changing the underlying interaction. Joint surface-tool centering removes sensitivity to common-origin translations, but cannot compensate for incorrect contact geometry or deviations in executed actions. The decision experiments also reveal that improved predictive accuracy and action ranking do not always translate into lower action-selection regret. The broader contribution of this work is to connect action representation with the physical conditions under which learned experience can be transferred and the information required to predict its consequences. This provides a basis for evaluating and developing object-centric and equivariant world models according to the physical relationships they must preserve, rather than geometric structure alone. Extending the analysis to measured contact states, more realistic material interactions, and closed-loop manipulation will help determine how these principles apply to a wider range of robotic tasks.

## Appendix A. Implementation and Reproducibility Details

The procedural generator and action sampler use the ranges in Table A.1. The core unified experiment series and extended mechanism tests use separate, fully recorded fruit, sample, noise and training seeds. The latter experiments vary frame translation, frame rotation, estimated contact, patch selection and actual executed action as independent controls rather than interpreting all deviations as a single noise parameter.

Table A.1. Geometry and action parameters used by the procedural peeling simulator. The reported ranges describe the main generator; individual controlled experiments restrict selected parameters.

| Parameter | Value used in main generator | Role |
|---|---|---|
| Nominal fruit radius | 40 mm | Global geometric scale |
| Main mesh | 642 vertices, 1280 faces | Learning experiments |
| High-resolution mesh | 2562 vertices, 5120 faces | Rigid-transform sanity check |
| Nominal peel thickness | 1.5 mm | Local physical state |
| Thickness clip | 0.90-2.175 mm | Procedural range before experiment-specific rescaling |
| Stroke length | 24-50 mm | Action variation |
| Penetration | 0.65-2.45 mm | Action variation |
| Blade inclination | -7 °to +7 ° | Action variation |
| Blade width | 7.5-12.5 mm | Action variation |
| Oracle path stations | 40 | Geometric consequence computation |
| Model blade-path samples | 8 | Predictor input |

The fruit generator uses ten low-order directional polynomial modes for shape and eight for thickness. With $\mathbf{u} = (x, y, z)$ as a unit direction in the fruit-body frame, $x$, $y$, and $z$ denote its dimensionless directional components rather than world-space position coordinates. The shape modes $\phi_1$ through $\phi_{10}$ are $z$, $(3z^2-1)/2$, $x^2-y^2$, $xy$, $xz$, $yz$, $z(5z^2-3)/2$, $x(5z^2-1)$, $y(5z^2-1)$, and $(x^2-y^2)z$. The thickness modes $\chi_1$ through $\chi_8$ are $z$, $(3z^2-1)/2$, $x$, $y$, $xy$, $xz$, $yz$, and $x^2-y^2$. These polynomial terms control smooth, reproducible departures from spherical geometry and uniform peel thickness. The radius is bounded below by $0.82r_0$.

The shape coefficients $c_1, \ldots, c_{10}$ are independent standard-normal draws scaled by 0.050, 0.035, 0.030, 0.026, 0.022, 0.020, 0.018, 0.016, 0.014, and 0.012. The thickness coefficients $b_1, \ldots, b_8$ use scales 0.10, 0.07, 0.06, 0.05, 0.04, 0.035, 0.03, and 0.025. Each random field is scaled by its nominal radius or thickness, as specified in Section 2.3. The Python implementation provides the exact random seeds, mesh construction, and normal calculations.

The matched PointNet-style predictor uses seven features for each sampled point: 3-D position, 3-D normal and peel thickness. Frame-noise and mechanism studies use 48 sampled patch points; the thickness and counterfactual decision experiments use 36. The action vector contains eight sampled blade positions plus blade width and angle. Frozen weights, standardization parameters, physical test tensors and seed-level predictions are retained for the mechanism ablations. The three new fruit cohorts each contain 14 fruit geometries, 700 training, 140 validation and 280 test interactions and five independently trained model seeds.

All representations use the same architecture and fixed physical scaling. Position and path coordinates are divided by 0.05 m. Thickness is centered at 1.5 mm and divided by 0.5 mm. Blade width is divided by 0.01 m. Blade angle is divided by 0.15 rad. Normals are left unitless. Outputs are standardized using training data. Models are optimized with AdamW at learning rate 0.002 and weight decay $10^{-5}$, mean-squared error, batch size 256, and early stopping on validation error. Maximum epochs and patience differ slightly between controlled tests because dataset sizes differ, but they are fixed before comparing representations within each test.

Global rigid-view augmentation transforms points, normals and blade paths while preserving Oracle labels. For the extended frame analysis, isotropic origin noise perturbs the estimated contact origin in all three axes, normal-only noise acts along the local surface normal, and tangent-only noise acts in its tangent plane. All share the unchanged true physical trajectory when frame noise is studied. The joint-centering control subtracts the sampled patch centroid from both patch positions and blade positions. Orientation tests rotate all local vectors together. Contact-localization tests either retain or reselect the sampled patch around the estimated contact. Execution tests perturb actual penetration/contact and recompute Oracle outcomes. The matched-update augmentation control retains the original 700 physical training samples per epoch and matches per-seed optimizer update counts across training-noise conditions; two independent fruit cohorts were evaluated. Complete per-model and per-noise tables accompany the reproducibility archive.

## Appendix B. Statistical Summary of Core Contrasts

The principal unified paired comparisons appear in Table B.1, with identical model-initialization seeds compared within each physical experiment. The exact two-sided Wilcoxon minimum with five nonzero pairs is 0.0625. It is descriptive rather than a significance gate. Table B.2 reports additional frame-noise crossover estimates across three independently generated fruit/sample cohorts. Those estimates are model- and noise-definition-dependent. The 15 fitted predictors are nested within three physical data cohorts and repeated test-noise draws do not constitute additional independently trained models.

Table B.1. Paired five-seed contrasts from the unified experiment series.

| Core contrast | Seeds | Mean A | Mean B | Mean paired difference | Seed-level direction | Two-sided Wilcoxon |
|---|---|---|---|---|---|---|
| Global relocation-World vs Object at $\rho$=1 | 5 | 1.0017 | 0.8584 | +0.1433 (World−Object) | 5/5 | 0.0625 |
| Equivalent surface relocation-Object vs Surface at 120° | 5 | 0.9250 | 0.8561 | +0.0689 (Object−Surface) | 5/5 | 0.0625 |
| Thickness withheld vs retained (Blind vs Full) | 5 | 0.8596 | 0.3888 | +0.4708 (Blind−Full) | 5/5 | 0.0625 |
| Accurate local frame-Surface vs Object | 5 | 0.5678 | 0.8951 | −0.3272 (Surface−Object) | 5/5 | 0.0625 |
| Counterfactual regret-Surface vs World | 5 | 0.2262 | 0.6528 | −0.4266 (Surface−World) | 5/5 | 0.0625 |

In Table B.1, means and paired differences are computed from the same frozen seed-level files that generate Figures 3,4 and 6. Error signs and Wilcoxon entries are shown for auditability rather than to imply population-level significance. The core contrasts retain a consistent paired direction across all five training seeds in the unified series. The principal mechanism conclusions are also independently supported by the algebraic cancellation of common translation, exact input collision when peel thickness is hidden, and distinct Oracle outcomes for executed-tool deviations. Additional response-curve statistics and their seed-level records are released alongside the figures to avoid inferring general sensor tolerances from small-sample crossing estimates.

Table B.2. Model-level crossover of Surface prediction NRMSE relative to a paired clean Object predictor under defined frame-origin perturbations.

| Perturbation type | Fitted models | Median first crossover | Crossing within range |
|---|---|---|---|
| Isotropic 3-D origin shift | 15 (3 cohorts × 5 seeds) | 0.840 mm | 15/15 by 12 mm |
| Origin shift along local normal | 15 (3 × 5) | 0.829 mm | 15/15 by 12 mm |
| Origin shift in tangent plane | 15 (3 × 5) | 6.829 mm, observed only | 13/15 by 12 mm |
| Shared systematic origin bias | 15 (3 × 5) | 1.318 mm | 15/15 by 12 mm |

In Table B.2, values are medians of individual model first-crossing estimates, linearly interpolated between tested $\sigma$ levels. Tangent-only statistics exclude two right-censored models. Each fitted model averages three noise draws. These noise definitions and cohorts differ from those used in the five-seed frame-boundary sweep in **Figure 4**(b).

## Author Contributions

Conceptualization, Y.S.; methodology, Y.S.; software, Y.S.; validation, Y.S.; formal analysis, Y.S.; investigation, Y.S.; visualization, Y.S.; writing—original draft preparation, Y.S.; writing—review and editing, Y.S. The author has read and agreed to the published version of the manuscript.

## Data Availability Statement

The numerical datasets supporting this study are provided in the accompanying supplementary reproducibility archive. The archive includes seed-level results, figure and table data for reproducing the plots and numerical summaries. These materials will also be uploaded to a public repository under the author's GitHub account (https://github.com/Yuhang-Sun).

## Conflicts of Interest

The author declares no conflicts of interest.